# Generative linguistics contribution to artificial intelligence: Where this contribution lies?

Mohammed Q. Shormani, Ibb University, University of Cyprus

**Abstract**
This article aims to characterize Generative linguistics (GL) contribution to artificial intelligence (AI), alluding to the debate among linguists and AI scientists on whether linguistics belongs to humanities or science. In this article, I will try not to be biased as a linguist, studying the phenomenon from an independent scientific perspective. The article walks the researcher/reader through the scientific theorems and rationales involved in AI which belong from GL, specifically the "Chomsky School". It, thus, provides good evidence from syntax, semantics, language faculty, Universal Grammar, computational system of human language, language acquisition, human brain, programming languages (e.g. Python), Large Language Models, and unbiased AI scientists that this contribution is huge, and that this contribution cannot be denied. It concludes that however the huge GL contribution to AI, there are still points of divergence including the nature and type of language input.

## 1.Introduction

Linguistics contribution to artificial intelligence (AI) has been underestimated by behaviorism fans; the debate centers around the "Chomsky School" and Behaviorism. If there is a contribution: how can this contribution be accounted for? If not, another question arises: where else has AI originated, or gained its ideas, concepts, and how? This study tries to prove that only the first alternative is valid. Some mathematicians and physicians with arrogance say that linguistics belongs to humanities and not science, without even looking at what linguists do, what topics they deal with, what science areas they involve, what theories they employ and propose (see e.g. Marcolli et al. 2023), in which the authors propose an algebraic approach to analyzing linguistic phenomena. In the final stages of writing this article, I have come across a strange statement by a Nobel physics winner, 2024, Geoffrey Hinton (GH), critiquing the "Chomsky School", a school whose age is about a century now with its extraordinarily vibrant achievements in linguistics, biology, psychology, cognitive science, among others. In his interview, GH said that not only the "Chomsky School" belongs to humanities, but also it is wrong. GH's statement is just a continuum of rumor by behaviorists since Chomsky's (1959) severe criticism of Skinner's thoughts. Chomsky along with generative linguists believes that behaviorist ideas do not suit human beings or are capable of describing their behavior. Given the huge developments of AI, Chomsky, also continuing his dissatisfaction, argued that whatever status AI reaches, it will never reach the human mind State or works like it (Katz 2012; Chomsky et al. 2023). For the "Chomsky School", the human brain/mind is perfect, a species-specific property, having an unprecedented working mechanism no other creature living on the planet ever possesses.

Linguistics, the scientific study of language, could be considered the "veins and nerves" of AI, specifically (neuro-)symbolic AI (see e.g. McShane and Nirenburg 2021; Maruyama 2021). The latter depends heavily on linguistic formal theory, having "a regular, context-free, and formal grammar" (Beysolow 2018: 2). It follows that linguistics and AI are correlated. This correlation is manifested through four factors (McShane and Nirenburg 2021: 20-40). First, linguistics gives rise to developing natural language understanding and generation capabilities within an integrated, comprehensive agent architecture (p. 20). Second, AI uses *human inspired, explanatory modeling techniques and actionability judgments* (p. 22). Third, it leverages *insights from linguistic scholarship and, in turn, contributing to that scholarship* (p. 34). Fourth, it incorporates *all available heuristic evidence when extracting and representing the meaning of language inputs* (p. 40, emphasis in the original).

Artificial intelligence originated from scientific questions as to how intelligence works, how our brain gives us cognitive abilities we possess, and how it can be implemented in computer (Katz 2012). AI dates back to the 1950s, perhaps with Isaac Asimov's 1942 seminal article 'Runaround', which was published in *Science Fiction Magazine* (Akhtar et al. 2022). Since its inception, AI has witnessed huge developments, the basic idea of which is "Can Machine Think?" (Turing 1950), stimulating computer programs to work, do tasks, think, and respond to human commands (Beysolow 2018). AI present state-of-the-art is developing Large Language Models (LLMs), based on Deep Neural Networks (DNNs), and Neural Networking Algorithms (NNAs). DNNs and NNAs have been utilized in several and various AI applications including NLP, machine translation, LLMs including ChatGPT and BERT (=Bidirectional Encoder Representations from Transformers) (Edunov et al. 2018; Linzen and Baroni 2021). Among all these, there has been a role played by the "Chomsky School". It has revolutionized human science and knowledge in several fields including "psychology, philosophy, cognitive science, and even computer science." (Golumbia 2009: 31). Don Knuth, an AI scientist, arrogantly says "I must admit to taking a copy of Noam Chomsky's *Syntactic Structures* along with me on my honeymoon in 1961" (Knuth 2003: x).

In this study, we try to understand how linguistics contributes to AI. We focus on three major questions: i) What is generative linguistics (GL)?, ii) How does AI develop form GL's ideas, notions, and hypotheses?, and iii) where do both areas converge and diverge? Thus, the article is organized as follows. Section 2 articulates the theoretical background of generative linguistics, demarcating several related issues including syntax, Universal Grammar (UG), L1 and L2 acquisition, and semantics. Section 3 spells out AI inception and developments, focusing on its present state, and its models including LLMs, their DNNs, and NNAs. This section also showcases LLMs' language acquisition, processing and interpreting of human language. Section 4 concludes the study.

**2. Generative linguistics**

Linguistics, the scientific study of language, is divided into several modules including phonology, morphology, syntax, semantics, and discourse. The central question of linguistics theorization is centered around language acquisition. Language is defined as "a set (finite or infinite) of sentences, each finite in length and constructed out of a finite set of elements" (Chomsky 1957: 2), as rules governing such constructions. Language, Chomsky argues, consists of a set of finite words and rules from which humans can generate an infinite number of pieces of language, be they phrases, clauses or sentences. To Chomsky, language's main/only function is not communication, as indicated by certain properties of language like ambiguity that humans "communicate and say things that turn out to be simple and fall short to express what we actually want." He sees language as a verbal (and nonverbal) system, rule-governed, complex but precise and concise. As for verbal system, he defines language as a verbal (phonology) system consisting of words (morphology), which carry meaning (semantics), structured in some way (syntax) in a particular situation (discourse). However, we will be concerned only with syntax and semantics.

## 2.1. Syntax

Generative syntax begins with Chomsky's (1957) seminal work *Syntactic Structures*. In this work, Chomsky plants the seeds of generative grammar. Generative linguistics aims to construct "a grammar that can be viewed as a device of some sort for producing the sentences of the [human] language" Chomsky (1957: 11). However, the field has passed through several developments and several phases. The first phase was the Standard Theory (ST) in which Chomsky proposes Kernel sentences. A Kernel sentence consists of an irreducible set of simple structures generated by Phrase Structure Rules (PSRs). From a Kernel sentence, different types of sentences can be formed by applying a number of Transformational Rules (TRs). This is exemplified in (1) and (2).

(1) a. Alia- story- read- a- has.
   b. Alia has read a story.

(2) a. Alia has read a story.
   b. Alia has not read a story.
   c. Has Alia read a story?
   d. A story has been read by Alia.

In (2c& d), there are two syntactic Transformation Rules (TRs) required by Yes/No question in (2b), and passivization in (2d), respectively. The second phase was the Modified ST. This was actually introduced in Chomsky's (1965) *Aspects of the Theory of Syntax*, thus incorporating DS and SS levels of representation into the syntactic theory. The third phase was the Extended Standard Theory (EST). The change in conception of EST is ascribed to the contributions of Jackendoff (1977) with his postulations of X-bar (or $X^I$) theory which adds to the exiting apparatus of syntactic theory the assumption that there are three levels in X projection: maximal projection, intermediate projection and minimal projection, XP, $X^I$, and $X^O$, respectively. It also adds binary branching hypothesis, which states that only two branches can come out of a node. These developments led to establishing the Principles and Parameters Framework (P&P), which marks a remarkable shift in the Transformational Generative Grammar (TGG) theorization, viewing language as a "multi-strata phenomenon" "housing" complex transformational mental operations between an invisible DS representation and a visible SS one (Veale 2006).

In P&P, there have arisen several questions as to what language is, how it evolves, how it is acquired, among others. In this framework, the study of mind and language and its biological bases marked the "second cognitive revolution" in the last eight decades or so. The biolinguistic approach to the study of language has declared that the knowledge of language humans possess is ascribed to a faculty in the brain called *Language Faculty* (FL) every human is genetically, innately and biologically endowed with (see e.g. Chomsky 1959, *et seq*; Cook 1983; Jenkins 2000; Hornstein 2009; Fodor and Piattelli-Palmarini 2010; Stroik and Putnam 2013).

The most notable shift witnessed in P&P framework has been the introduction of biology into the syntactic inquiry, making it clear that humans have a faculty of language which is the responsible part for language acquisition, computation, and perception of language. FL contains Computational System (of human language) ($C_{HL}$) and UG. The latter contains a set of rules called *Principles* and *Parameters*. *Principles* are shared by all the languages of the world, while *Parameters* are language-specific. For example, the *Subject Principle* states that "every clause must have a subject" and this is present in all languages of the world, while the *Subject Parameter* states that "the subject is either overt or covert (implicit)." The first case is found in languages such as English, Hindi and French, and the second case in languages such as Arabic, Greek and Irish. All these developments led to Minimalism.

Minimalism is the latest development of syntax theory, a theory whose major concern is that to study language, linguistic theory should use few primitive machinery apparatuses. Crucial to Minimalism is what has been called by Chomsky (2000: 97) the Strong Minimalist Thesis (SMT) which is formulized in (3).

(3)   Language is an optimal solution to legibility conditions

In Chomsky's view, the notion of "legibility conditions" is related to interface properties. Put differently, the idea tied to legibility conditions is that the core $C_{HL}$ (or otherwise, the narrow syntax) makes available the optimal way of relating an arbitrary set of lexical items to the interfaces, i.e. FL (i.e. the conceptual-intentional system) and PF (the conceptual-intentional system). (3) could be understood in such a way that language is an optimal solution of the $C_{HL}$ to the constraints imposed by the two interfaces. These interfaces are related in such a way that *meaning* is tied to *sound* to satisfy whatever conditions imposed by the *intrinsic properties* of lexical items and the interfaces. The Minimalist Program seeks an adequate linguistic theory to describe what is so called *I-language*, i.e. UG (or the *Initial State*) of "an ideal speaker-listener in a completely homogenous speech community who knows its language perfectly" (Chomsky 1965: 3) and not his *E-language*, i.e. what he says, (performance). Chomsky (2013: 36) maintains that the fact that "[e]mbedding the study of *I-language* in the biolinguistic framework is entirely natural; an individual's language capacity is, after all, an internal property" still persists. However, UG, as an internal and integrated part of human FL, and in its "technical sense" should "not be confused with descriptive generalizations" like those advocated in Greenberg's universals.

**2.1.1. Universal Grammar and language acquisition**
Recall that the biological theory has been considered a theory that tries to characterize the innate human 'language faculty,' which is viewed as any other performance modular system in terms of its behavior. Biologically, language faculty is considered to be an "organ of the body," along with other cognitive systems. The idea of the biological nature of the human faculty of language comes from linguistics. Following Chomsky's early ideas, some linguists (see e.g. Jenkins 2000; Stroik and Putnam 2013) hold that linguistics suggests "core internal properties of the language faculty, that in turn posed important questions for biology," and that the biological properties are related to UG, and "the syntactic computations of the language faculty are the biological evidence" (Jenkins 2000: 3).

Furthermore, FL is sometimes equated with language (Chomsky 2005). Biolinguistics in general views a person's language as a state of some component of the mind/brain. Following the common assumption in biolinguistic approaches to language study, Chomsky (2000, *et seq*) holds that language faculty resembles or is part of human intellectual capacity, arguing that "whatever the human intellectual capacity is, the faculty of language is essential to it" (Chomsky 2005: 3). If the faculty of language, as assumed by Chomsky, possesses "the general properties of other biological systems, we should, therefore, be seeking three factors that enter into the growth of language in the individual:

  i. Genetic endowment, apparently nearly uniform for the species, which interprets part of the environment as linguistic experience, a nontrivial task that the infant carries out reflexively, and which determines the general course of the development of the language faculty. Among the genetic elements, some may impose computational limitations that disappear in a regular way through genetically timed maturation.

 ii. Experience, which leads to variation, within a fairly narrow range, as in the case of other subsystems of the human capacity and the organism generally.

iii. Principles not specific to the faculty of language.

These three factors, Chomsky argues, are the cornerstones of not only the language faculty but also of language itself. In (i), for instance, genetic endowment of language is species-specific. This genetic endowment may also contain the computational system which is responsible for computing the linguistic objects, i.e. lexical items *selected* from the lexicon. (ii) seems to suggest a relation between the language faculty and other systems of human capacity. The third factor, Chomsky holds, "falls into several subtypes: (a) principles of data analysis that might be used in language acquisition and other domains; (b) principles of structural architecture and developmental constraints that enter into canalization, organic form…including principles of efficient computation" (Chomsky 2005: 6).

The above three factors, specifically the third one, provide a plausible answer to "[t]he general question … How far can we progress in showing that all language-specific technology is reducible to principled explanation, thus isolating the core properties that are essential to the language faculty, a basic problem of linguistics?" (Chomsky 2005: 11). In his own words: "language independent principles of data processing, structural architecture, and computational efficiency… [answer] the fundamental questions of biology of language, its nature, use, and perhaps even its evolution" (Chomsky 2005: 9). Along these lines, Stroik and Putnam (2013: 6) argue that "the evolution of organisms is not primarily governed by extrinsic forces, but rather is driven by the design properties internal to the organism in question."

The fact that young children easily acquire any language spoken in the environment around them has been viewed as a miracle (Jenkins 2000). From a generative perspective, "this is only explicable if certain deep, universal features of this competence are innate characteristics of the human brain." If this is true, then, it must be that this ability or the "inheritable capability to learn any language means that it must somehow be encoded in the DNA of our chromosomes" (Jenkins 2000: 4). This, in principle, explicitly implies the universality of language. However, a question arises here: what is this universality of language all about? The above argument in a way or another gives us enough room to postulate that children acquire language because they are endowed/predisposed with mental ability wired in their genes, and what the environment does is just provide them with linguistic input necessary for language acquisition to take place. When language acquisition takes place, be it that of L1 or L2, a child or adult, respectively, possesses a perfect system. At the age of 2-3 years, a child may have been exposed to millions of tons of phrases and sentences. A child, say, at the age of 3-5 years, is able to produce a language similar to that of adults. S/he even is able to produce pieces of language s/he has never heard or come across before.

In the case of L2 acquisition process, adult learners may have been exposed to millions of tons of phrases and sentence in or out of classroom. L2 acquisition is simply defined as acquiring a language subsequent to L1. From a UG perspective, L2 acquisition is resetting UG parameters already acquired along with L1, given that UG principles are not acquired in L2. Consider (4) spoken by a freshman student (Shormani 2014, 2016).

(4) What I want to say is that who comes early to class should sit in the front bench, either short or tall.

The sentence in (4) is complex in the technical sense, and perhaps the student has never heard or come across before. It consists of a matrix clause and embedded clauses. The wh-word *what* is the object of the verb *say*, but it is fronted, i.e. it occurs before the subject, viz. the pronoun I. The embedded clause *who comes early to* class is the subject of the verb *should*. In addition to all this, the student in question may not have come across this sentence before, simply because s/he has not studied syntax, poetry, novels, plays, etc. in which we may think that s/he may have come across similar constructions. The question imposed above, thus, repeats itself: how can the student in question produce (4)? It seems difficult (and perhaps impossible) to answer this question without relating (4) to UG. Put differently, the student in question cannot produce (4) unless the UG principles concerning wh-fronting, topicalization, embedding, phrase structure rules, etc. have been activated previously, namely by the student's L1. Consider (5) which is almost a similar Arabic structure to (4).

(5) ʔalli ʔašti ʔaquuluh ʔinnuh man yiji mubakir laazim yajlis fii ʔawal ṣaf, sawaa (kaan) qassiir ʔaw ṭawiil

(5) shows us that what the student has to do is just *reset* UG parameters s/he has acquired while acquiring his/her native language, i.e. Arabic, given the fact that UG principles have been activated by L1 Arabic.

The assumption that the student has not heard, or come across, but is able to produce structures like (4) perhaps reveals the "poverty of the stimulus", given that the setting is foreign, a university hall. Put differently, the English spoken around him/her is "poor," which may not be enough to enable him/her to say (4). This means that the student "generates" a sentence in L2, i.e. English, from the "poor" linguistic input s/he has been exposed to, but it is enough in the sense that it enables him/her to "generate" (4). This is perhaps akin to the question asked by Chomsky in the case of child language acquisition, i.e. how difficult structures are acquired and perhaps judged by a child in an early age when his/her brain is not matured enough to cope with abstract concepts. This also supports the claim that it is UG which compensates for the "poverty of the stimulus" and explains pieces of language occurring in the child's language, though s/he has never heard or come across before.

## 2.1.3. Computational system of Human language

The C$_{HL}$ has several mental operations, the most important of which are four: first, the *Select* operation selects a LI only once and hence, meeting the *Inclusiveness Condition* (IC) which prevents other new elements and/or features to be introduced into the computational system once more. Examples of this is the nouns' [-/+Def], Case and φ-features that are assigned in the numeration either as intrinsic features by lexical entry or optional features by the numeration operation. The second operation is called *Merge*. This operation merges two items (functional and/or lexical) (α, β) forming K(α, β) asymmetrically projecting into a single item, either α or β where the projecting item becomes the head, and hence, the label of the resultant complex, i.e. a phrase (Chomsky 2013), and *Merge* is recursive, i.e. it takes place as many times as necessitated by the derivation and the nature of the piece of language under consideration. The third operation of C$_{HL}$ is *Agree* which establishes a relation between a lexical item *a* and a feature *F* in some restricted Search Space (Chomsky 2001). This relation is then manifested in what is so called *agreement*. For instance, Case-checking/valuation is an *Agree* relation established between a lexical object say K labeled LB(K) and a feature F in some restricted search space (i.e. the K-F's c-command domain in either directions) (Chomsky 2000, 2001) where each is φ-complete in order for the valuation process to take place. Since the LB(K) is the only element of K that is immediately accessible to a language L, it has to be the element that activates *Agree*, by virtue of its unvalued features which constitute a *Probe* that seeks a matching *Goal* within the domain of LB(K). The relation *Match* is taken to be an identity (Chomsky 2001: 4). Matching of probe-goal induces *Agree*, eliminating unvalued features that activate them. The fourth operation is *Spell-out*, which gives a piece of language its phonetic form.

(6) C$_{HL}$

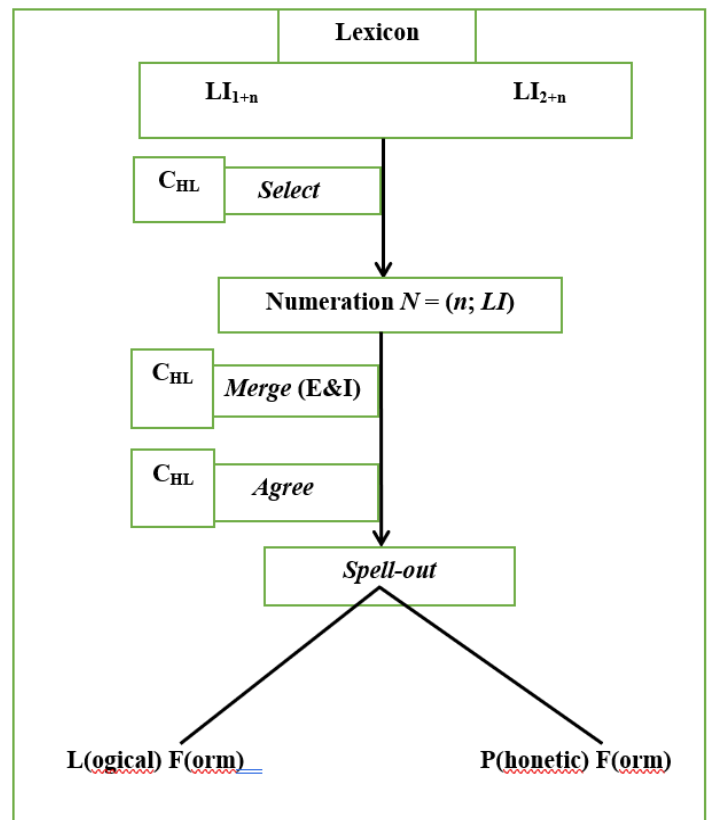

**Fig 1: Computational system of human language**

Consider (7) and (8). (8) is the feature composition of T and *Ali*, the *probe* and *goal*, respectively.

(7) Ali eats an apple.

(8) T [*u*EPP; *uφ*; *v*Nom]; Ali [*v*EPP; *vφ*, *u*Nom]]

After all these operations are done, the phrase/sentence is spelled-out through the *Spell-out* operation. It is then sent to the interfaces, viz., LF and PF.

Let us now concretize our argument, answering the question: how does human brain process structures, be they simple, or complex? Consider how (7/9) is derived, as schematized in (10).

(9)    Ali eats an apple.

(10)

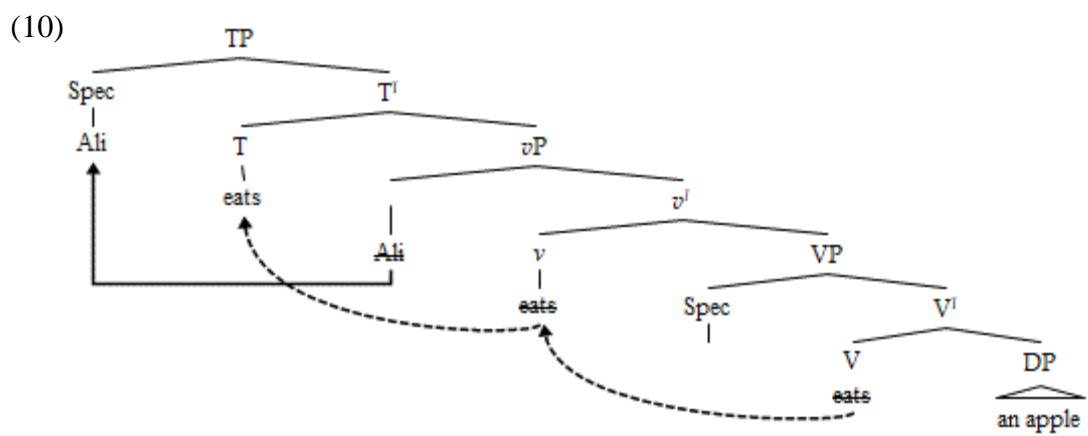

The human brain processes a sentence like (9/10) as follows. The operation *Select* selects the noun *apple* and the Det *an* which are LIs from the lexicon/numeration in pairs, based on their selectional restrictions, i.e. in this case only *an* can collocate with *apple*, as we will see shortly. The operation *Merge* then comes to play. Both constituents will form a pair under DP projection forming the pair DP{an; apple}. Another *Merge* takes place merging the verb *eats* with the DP constituting V$^I$. Given X$^I$ schemata, the V$^I$ merges with its Spec, forming VP. The VP projection requires a *v*P (VP-Shell hypothesis, see Larson 1988). Another *Merge* merges the DP *Ali* in Spec,*v*P, as the subject of the sentence. Merging in Spec,*v*P, the DP *Ali* receives the θ-role of *Agent*, from *v*, which in turn values the θ-role of *Patient*. The structure in (12) is finite, which necessitates merging a functional projection, viz., TP. Given this almost complete derivation of the structure in (9), the verb *eats*, though selected from the lexicon inflected for all φ-features, needs to check/acquire tense, which in turn would necessitate it to move to T through *v* in what is so-called Head movement (see e.g. Travis 1984). In this point of derivation, *Agree* comes to play. *Agree* will take place between T and *Ali*, the probe and goal, respectively, the result of which is valuing and interpreting all the unvalued and uninterpretable features of T and Ali including φ-features, and Nom case. Once these features have been valued and interpreted, they are deleted immediately. Given that T has an EPP feature, which is satisfied/valued by (re)merging a DP in its Spec. Given that in T's Search Space, the nearest DP is *Ali*, the latter has to move to Spec,TP via *Internal Merge* (IM).

Given that UG principles are universal, and UG is parameterized cross-linguistically, how does human brain process (9/10)? One such parameterization between English and Arabic, for example, concerns word order, i.e. while English is an SVO language, Arabic is a VSO language. Consider (11) which is an equivalent Arabic structure to (9).

(11) yaʔkulu ʕaliyyun tuffaħatan
    eats     Ali     an.apple
  'Ali eats an apple.'

In computing/deriving (11), there is only one difference between (9) and (11), i.e. while *Ali* undergoes an IM from Spec,*v*P to Spec,TP in English, it does not do so in Arabic. Thus, the human brain processes/derives (9) and (11) almost similarly. However, there is still a problem, i.e. how is T's EPP feature valued?, given that ʕaliyyun remains in-situ. This causes the derivation to crash at LF. This issue has been discussed thoroughly in the literature concluding that since languages like Arabic are rich inflected languages, T's EPP feature in these languages is valued by this richness of inflection (see e.g. Alexiadou and Anagnostopoulou 1998; Koeneman and Zeijlstra 2014; Shormani 2015a).

Another parametric variation between Arabic and English concerns *pro*-drop property. To concretize our argument, consider (12).

(12) yaʔkulu *pro* tuffaħatan
    eats         an.apple
  'He eats an apple.'

In (12), the subject seems to be covert, i.e. unpronounceable; human brain processes (12) considering *pro* a subject. It follows that (12) will be processed by the human brain as showcased in (13).

(13)
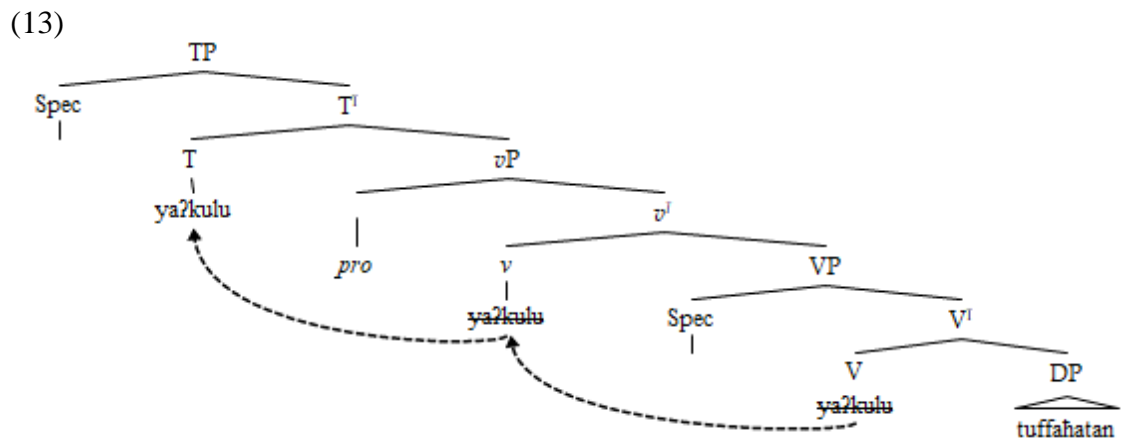

The derivation processes in (10) will take place in (13), too. The only difference is that while the subject in (9/10) is an overt DP and raises to Spec,TP, it is *pro* in (13), and remains in Spec,*v*P.

Another aspect to tackle here is long-distance agreement (LAD), and see how human brain processes complex structures like (14).

(14) What has Ali thought that Alia will like in level four?

(14) is a complex wh-question whose complexity lies in the fact that it contains multiple movements of the wh-word *what*. (15) below is the derivation of (14), see how the human brain processes such complex structures.

(15)

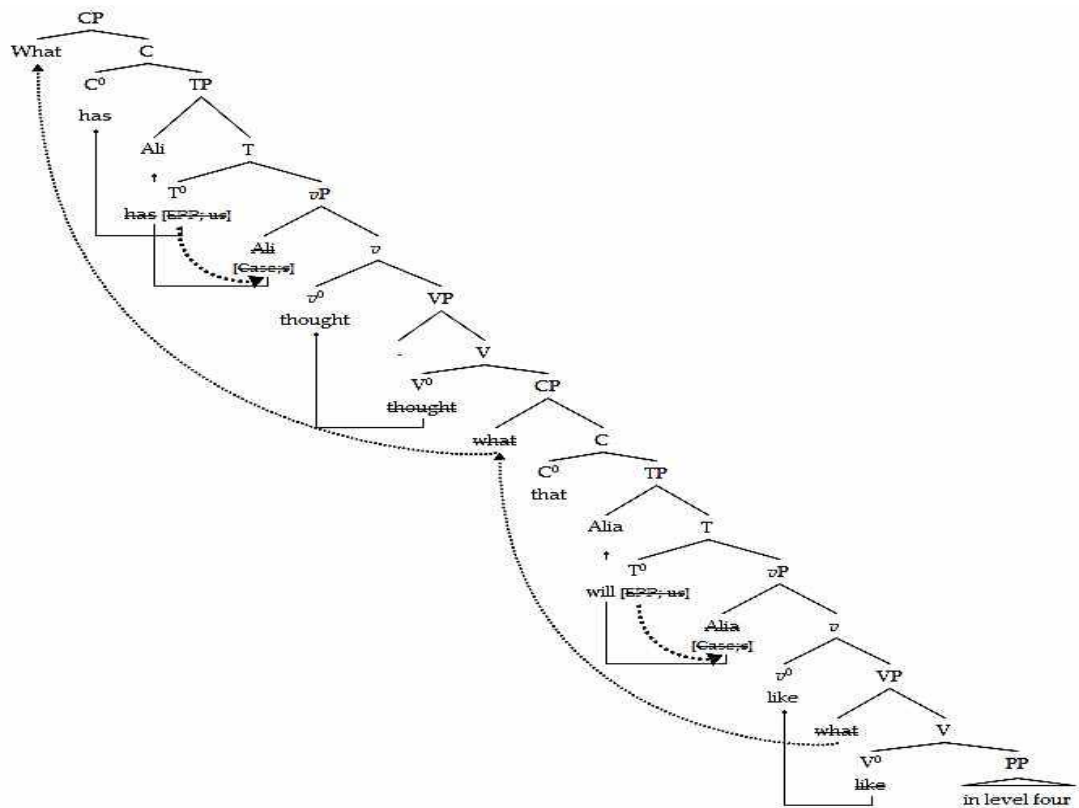

In (15), the verb *like* will raise to $v^o$ where it enables it to assign θ-role to its subject DP *Alia*. Being the probe, $T^o$ is active (by virtue of having unvalued features like EPP, $u\varphi$), it enters in an *Agree* operation with the subject *Alia*. The DP-subject *Alia* raises to Spec,TP by virtue of being triggered by T's EPP. The derivation then proceeds in such a way that the CP merges with $V^o$ projecting into VP, which in turn merges with $v^o$ and hence projecting into *v*P. The process iterates to reaching the CP. Note that the wh-word *what* is still in situ. Another thing required by wh-question formation and it is still not performed is that there is no Subject-Aux inversion. The Aux of the matrix clause undergoes a raising movement to $C^o$. The wh-word *what* first raises to Spec,CP (i.e. the lower one. From there, it raises to Spec,CP (i.e. the higher one). Note that the canonical position of *what* is V-Comp configuration, but it is merged in Spec,VP, just for easy exposition.

## 2.2. Semantics

Semantics is simply the study of meaning. To linguists, semantics is another component of the grammar. We will focus only on selectional restrictions, lexicon and argument structure.

### 2.2.1. Selectional restrictions

A lexical item, say, a verb has, in addition to subcategorization restrictions, some semantic restrictions which are essential "for computing semantic relationships between elements in a sentence" (Myers and Blumstein 2005: 279). If those restrictions are violated, such relationships will not be computed. Thus, while subcategorization restrictions are concerned with the number of the categorial types or c-selection (Category Selection, i.e. DP, AP, PP, CP) of the constituents that occur to the right of a lexical head, selectional restrictions are concerned with the S-selection (Semantic Selection) of such constituents in both sides of the predicate. The former are referred to as the ability of the head to restrict the number and type of the internal arguments and the latter as the ability of the head to restrict the type of both internal and external arguments which can be assigned θ-roles. Put differently, selectional restrictions restrict the categorial features of the constituent that can occur not only to the right of a particular lexical head but also the one(s) occurring to its left. Subcategorization restrictions do not state anything regarding the specific type of lexical items that can function as complements of verbs, for instance. Selectional constraints (restrictions) are to be imposed on the lexical item to be able to function as a subject/complement, or otherwise as an argument in general. Subcategorization restrictions are syntactic in nature, while Selectional restrictions are semantics-based.

Thus, from a P&P perspective, the above argument gives us a clue that the relation between UG and lexicon is more or less manifested in subcategorization restrictions where a lexical category, be it a verb, noun, adjective, etc. subcategorizes for a complement. However, subcategorization restrictions do not specify what goes with what in a phrase. As noted above, this could be compensated by selectional restrictions which are part of semantics. These selectional restrictions are represented by categorial features encoded on the lexical items. These categorial features are manifested in terms if inflections.

### 2.2.2. Lexicon: feature specification

In P&P, selectional restrictions are said to be represented by categorial features encoded on a lexical item. In Minimalism, categorial features are expressed by means of *feature specification*, i.e. every lexical item is specified for all the features it carries in the lexicon. Feature specification is binary in nature, i.e. features are specified in a binary fashion, i.e. [+/-]. For instance, the noun *book* is [+N, +inanimate, -abstract]. The noun *honesty* is [+N, +inanimate, +Abstract]. Nouns are specified as [+N, +A, -V]. Nouns can take adjectives as modifiers but not verbs. Verbs are specified as [+V, +Adv, -A], in the sense that verbs are modified by adverbs but not nouns or adjectives. Verbs, in case of arguments, are specified as [+V, +/- Trns], in the sense that verbs can be transitive or intransitive. The same is also applied to other lexical categories. But then, a question arises, i.e. does it mean that this is left open by UG? If UG has nothing to do with feature specification, it follows that a native speaker of a language, say, English, will commit errors in collocation, which is not the case. Thus, it seems that UG, in addition to providing us with subcategorization rules, provides us with some kind of "selectablity," which makes us, as native speakers of an L, able to "select," as a priori, the context that "fits" a particular lexis to co-occur with. For instance, native speakers of English select the adjective *strong* to collocate with *tea* rather than the adjective *powerful* to collocate with it (Shormani 2015b).

Let's concertize our argument by considering structures like (16):

(16) a. nice weather

    b. *kind weather

In (16a), there is collocation in which the adjective *nice* collocates with the noun *weather* and the result is a grammatical collocation. However, the collocation in (16b) is ungrammatical (Shormani 2015b).

**2.2.3. Argument structure**

In this section, we will briefly introduce argument structure. It is typically a semantic notion which refers to the number of lexical items a predicate can take. It has been adopted from mathematics, in which a function can have one or more arguments. A predicate can be any lexical category including verb, noun, and adjective. In semantics, predicates can be divided into four types: a zero-place or zero-argument predicate as in the case of *rain* in *it rains*, one-place predicate like *laugh*, two-place predicate such as *kill*, and three-place predicate including *give*. An argument can carry a θ-role of *Agent*, *Patient, Theme, Goal*, depending on the predicate, here the verb, they cooccur with. For example, the argument structure of (9) is that the verb *eat* is a two-place predicate, *Ali* is the external argument and *an apple* is the internal argument.

**3. Artificial intelligence**

The real beginning of AI is attributed to Isaac Asimov's 1942 short tale 'Runaround'. This article was published in 'Science Fiction' magazine. In this seminal text, Isaac Asimov formalized his three laws of robotics: i) robots are not harmful to humans, ii) a robot obeys human commands, and iii) a robot must defend itself (Haenlein and Kaplan, 2019; Akhtar et al., 2022). AI's first definition was "making a machine behave in ways that would be called intelligent if a human were so behaving" (McCarthy et al., 1955: 11). In the early 1960s, there was an International Conference held at MIT that gathered linguists, philosophers and computer scientists worldwide including John McCarthy, Marvin Minsky, Nathaniel Rochester, and Claude Shannon the leading figures of AI with GL background or interest (Kibbee 2010). The most notable conclusions and recommendations of this conference were that they should "understand what intelligence is and how it can be put in computers". They found that there is a strong relation between human intelligence and (structure of) language, because "[l]anguage is one of the most complex and unique of human activities, and understanding its structure may lead to a better theory of how our minds work" (Winograd 1971: 15). There is, therefore, a relation between how human brain works and intelligence of the one hand, and between these and language.

Newell and Simon's research to model human reasoning by computer was inspired by Chomsky's ideas and argument of formal language in his paper entitled *Three models in the description of language* (Kibbee 2010). The correlation between AI and GL was very strong represented by Newell and Simon's (1956) paper entitled *The logic theory machine - a complex information processing system* and Chomsky's (1956) paper, just mentioned. Newell and Simon, AI specialists, were very much impressed with Chomsky's formal arguments about language, both considering these arguments to meet their own ideas. For example, describing this correlation, Simon (1996: 75) characterized the two papers, and the research programs they represent, as sharing a deep intellectual bond (see also Kibbee 2010):

> Historically the modern theory of transformational linguistics and the information-processing theory of cognition were born in the same matrix — the matrix of ideas produced by the development of the modern digital computer, and in the realization that, though the computer was embodied in hardware, its soul was a program One of the initial professional papers on transformational linguistics and one of the initial professional papers on information-processing psychology were presented, the one after the other, at a meeting at M.I.T., in September 1956 [names of the papers]. Thus the two bodies of theory have had cordial relations from an early date, and quite rightly, for they rest conceptually on the same view of the human mind (Simon 1996: 75).

In this excerpt, we find that Chomsky's paper has made great impact not only on Newell and Simon's research and ideas, but also on all AI academic scholarship. It paved the road for AI specialists and theorists, and perhaps until recently even if not confessed by GH. In his (1956) paper, Chomsky states:

[W]e picture a language as having a small, possibly finite kernel of basic sentences with phrase structure in the sense of §3, along with a set of transformations which can be applied to kernel sentences or to earlier transforms to produce new and more complicated sentences from elementary components. We have seen certain indications that this approach may enable us to reduce the immense complexity of actual language to manageable proportions and; in addition, that it may provide considerable insight into the actual use and understanding of language (Chomsky 1956: 124).

Since its inception, AI has witnessed huge developments, the basic idea of which is to stimulate computer programs to think, work, do tasks, and respond to human prompts. In 1966, the ELIZA program, an NLP tool, was invented by Joseph Weizenbaum. This NLP tool had the ability to converse with human, eventually known as AI and heuristic programming (Weizenbaum 1966). In the 1988s, Neural Networks came to play, and developed fast through research backpropagation (Zhang and Lu 2021; Akhtar et al. 2022). According to Delcker (2018), driverless vehicles like Mercedes-Benz we developed, having cameras and sensors with computer system which controls the steering (see also Akhtar et al. 2022). The most notable output of the work perhaps includes the fact that if they want to put human intelligence in computers, they should find a means. If we ask ourselves: how do computers work?, we will find that computers consist of hardware and software. The former is the hard components including the motherboard. However, software are programs that enable the computer to function. But again, how are these programs built? These programs are built or constructed by means of programming languages such as C++ or Python. These languages are nonverbal systems. A nonverbal language could be simply defined as a nonverbal (no phonology) system consisting of words/signs (morphology) whereby meaning (semantics) is delivered in a particular structure (syntax).

**3.1. Programming language**

(18) CLg:

| aaaa | aaab | aabb | abbb |
| bbbb | bbbc | bbcc | bccc |
| cccc | cccd | ccdd | cddd |
| dddd | ddde | ddee | deee |
| eeee | eeef | eeff | efff … |

As has been stated so far, the very idea of languages like (18) comes from Chomsky (1956) and automata theory. The latter studies abstract machines and computational problems and how automata can be used to solve such problems (cf. Beysolow 2018).

Chomsky in his seminal paper (1956) puts the first seeds of programming languages, and with Newell and Simon (1956) the first seeds of AI, in general (see also Chomsky, and Miller 1958). For instance, in Chomsky (1956) he foresees what formal language is, and how it consists of infinite sets of sentences each having finite elements (pp. 113-114). He tries to conceptualize and formalize similar languages like (18) (pp. 117-118). The approach Chomsky put the seeds for is Generative Linguistics, which assumes that language is a rule-governed phenomenon, and human brain can generate an infinite set of sentences following such rules. As we have seen in section 2, human language faculty has a computational system which simply means that from a "set of finite words and rules… humans can generate an infinite number of pieces of language, be they phrases, clauses or sentences" (Shormani 2024: 2). Chomsky proposes that $C_{HL}$ consists of neurons/neural networks and algorithms which work in perfect mechanisms, reflecting the SMT, or the optimal design of the language faculty.

Let Python be the language in (18), then naturally it has syntax and semantics (at least due to being "nonverbal"). Consider the following program from Python:

```
(19)  >>> import nltk_lite.draw.plot
      >>> from nltk_lite.corpora import gutenberg
      >>> count = { }
      >>> for w in gutenberg.raw():
             if w not in count:
                 count[w] = 0
             count[w] += 1

      >>> freq = [(freq, word) for (word, freq) in count.items()]
      >>> freq.sort()
      >>> freq.reverse()
      >>> print freq[:50]
```

This program is simple and depends on WordNet corpus as the dataset. The program asked for the first 50 most frequent words in WordNet corpus, as indicated by [:50], the colon and the digit 50 in the command >>> print freq[:50]. The result of this program is given in (20):

(20) [(137902, ','), (91197, 'the'), (60549, 'and'), (53788, 'of'), (50402, '.'), (31634, 'to'), (22369, 'in'), (21253, 'I'), (19296, 'that'), (18917, 'a'), (18649, ';'), (15473, 'And'), (15441, ':'), (14628, 'he'), (13782, 'his'), (13209, 'not'), (12633, 'be'), (12549, 'it'), (12347, 'for'), (11892, 'with'), (11718, 'is'), (11408, 'was'), (10978, 'shall'), (9915, '"'), (9760, 'him'), (9553, 'all'), (9309, 'they'), (8949, 'unto'), (8894, 'you'), (8621, 'her'), (8584, 'them'), (8501, 'as'), (7674, 'my'), (7323, 'have'), (7297, 'me'), (6651, 'LORD'), (6507, 'had'), (6450, 'from'), (6447, 'which'), (6086, 'their'), (5966, '?'), (5960, 'but'), (5776, "'"), (5757, 'will'), (5687, 'thou'), (5660, 'said'), (5503, 'by'), (5429, 's'), (5372, 'on'), (5283, 'The')]

In (20), the most frequent word/element is ',' with freq of 137902. And the least frequent word/element is 'The'. Now, if we make the command >>> print freq[150], the result is (1409, 'It'). If we want only the most frequent word in the given corpus, viz. >>> print freq[50], the result is (5205, 'thy').

If we want to count the number of words in the corpus in terms of types and tokens, we will construct the following Python program:

(21)  >>> counts={}
      >>> g='token type'
      >>> from nltk_lite.corpora import gutenberg
      >>> for token in gutenberg.raw():
              for type in token.split():
          if word in counts:
              counts[word]+=1
          else:
              counts[word] = 1
       >>> for s in sorted(counts):
              if s in g.split():
          print counts[s], s

The result of this program is given in (22):

(22)  5614086 token
      3180341 type[1]

The fact that programming languages such as Python and C++ depend on syntax and semantics in constructing programs ensues from the fact that if some principle is violated, we will get a notification of either a syntax error or semantic one (see e.g. Jackson 2018). Let's first try to construct a Python program and violate a syntactic principle. Consider the following program:

(23)  >>> from nltk_lite.corpora import gutenberg
      >>> count = {}
      >>> for w in gutenberg.raw():
          if w not in count:
      ^       count[w] = 0
          count[w] += 1
      >>> SyntaxError: invalid syntax

The invalidity of syntax of this program ensues from space in the code "if w not in count:", as indicated by the caret (^) which points to the code having a problem. Additionally, the error message "SyntaxError: invalid syntax" helps us verify the type of the syntax error (see e.g. Možina and Lazar 2018; Jackson 2018). This line of command needs, say, a tab rightward as in (24):

(24)  >>> from nltk_lite.corpora import gutenberg
      >>> count = {}
      >>> for w in gutenberg.raw():
              if w not in count:
                  count[w] = 0
              count[w] += 1
      >>>

---

[1] Note that these programs are constructed in Python 2.4.3. Note also that the terms "token" and "type" appear in the program in their singular form.

After pressing "enter" we get the sign ">>>" which means that the program is syntactically correct. Now, we can write the code we want using the function print.

As for semantics, a semantic error stems from the fact that the meaning of the program is written inappropriately. For example, if we change the word "if" or "else" in (19) or (21) and use any other word, we will get a semantic error notification (see also Jackson 2018).

**3.2. Large Language Models**

**3.2.1. ChatGPT**

ChatGPT is perhaps the most powerful artificial intelligence LLM based on neural networks, launched by OpenAI in 2022 (Kung et al. 2023; Siu 2023). It is "a machine-learning system that autonomously learns from data and can produce sophisticated and seemingly intelligent writing after training on a massive data set of text" (van Dis et al. 2023: 224). It has been used in various disciplines including research, academic writing, education, medicine, among others, performing tasks such as chatting, summarizing, creating content, and completing codes (Ray 2023:122; see also Cascella et al. 2023; Sallam et al. 2023). ChatGPT was trained on gigantic amounts of data. After the training process, ChatGPT now can generate human-like responses and these responses are generated in situ, based on the abstract relationship between words ("tokens") in the neural network. and more importantly the academic sphere, and students and researchers' life, in particular (Dergaa et al. 2023; Lee 2023). Dergaa et al. (2023), for instance, argue that ChatGPT is able "to create well-written student essays, summarize research papers, answer questions well enough to pass medical exams, and generate helpful computer codes". They add that ChatGPT can also write "research abstracts that scientists found difficult to distinguish from those written by a human" (p. 615). It has several consequences far-reaching "for science and society" (van Dis et al. 2023: 224).

Like other DLL models including BERT, ChatGPT basically consists of several transformers called *transformer blocks*. It comprises "two deep neural networks, namely the encoder and decoder". After processing the source text, the encoder converts it into an internal representation, which is utilized by the decoder "to generate target text by predicting each subsequent target word" (Siu 2023: 2). As Fig 2 demonstrates, and from bottom-up, the first transformer is *Text and Position Embed*, and the second is *Masked Multi Self Attention*, and so on until reaching the *Text Prediction* and *Task Classifier* transformers.

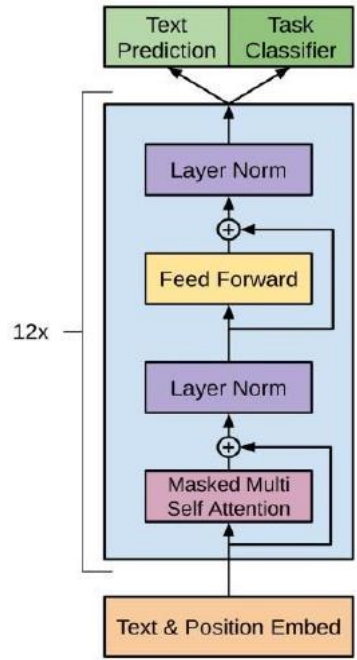

**Fig 2: ChatGPT Architecture (Vaswani et al., 2017).**

The working mechanism in which ChatGPT works makes it different from, say, other chatbots or conversational systems that are permitted to access external sources of information (e.g. performing online searches or accessing databases) in order to provide directed responses to user queries. But, it outperforms several other LLMs (Kung et al. 2023; Ray 2023).

### 3.2.2. BERT

BERT can be defined simply as a bidirectional LLM which encodes representations from transformers. It is a pre-trained deep language representation model, trained on huge text corpora. Traditionally, BERT was pre-trained in a left-to-right or right-to-left manner (Radford et al., 2018). The very idea of "bidirectional" is that BERT model can be trained either left-to-right or right-to-left, in a way the bidirectional conditioning allows the "word" to see "itself". However, it has recently been trained using two unsupervised tasks (Devlin et al., 2019). It was originally shown that BERT model's architecture is a multilayer bidirectional Transformer encoder (Vaswani et al., 2017). There are two important processes in BERT working mechanism: pre-training and fine-tuning. In the former.

There are several unlabeled data corpora used for training BERT. For example, Devlin et al. (2019) used the BooksCorpus (Zhu et al., 2015) and English Wikipedia in the pre-training process. The former contains 800M words and the latter 2500M words. In fine-tuning process, parameters are all fine-tuned using ImageNet, for instance. Fig 3 displays both processes.

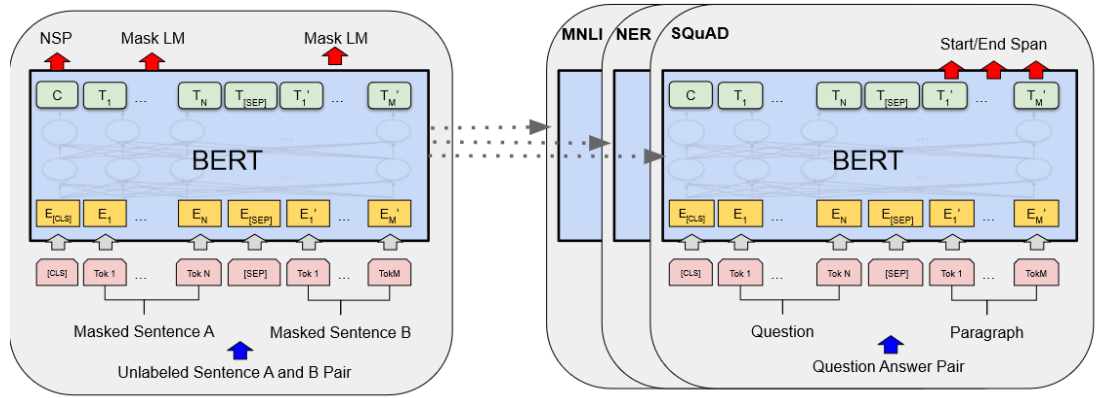

Fig 3: pre-training                fine-tuning (Devlin et al. 2019: no p.).

### 3.3. LLMs' language acquisition

Recall that the human brain has $C_{HL}$ which contains sets of rules, computation algorithms, and lexicon. FL contains UG, having principles and parameters, and can learn any natural language, UG principles are available in all natural languages, and UG parameters are language-specific (see e.g. Chomsky, 1981, *et seq*; Cook 1983; Long 2003; White 2003; Clark 2009; Shormani 2023). Incorporating Neural Network Algorithms enables LLMs to work intelligently like the human brain, say, having a human brain working mechanism. They are called Deep Language Learning Models (DLLMs), performing tasks with considerable accuracy (Gulordava et al. 2018; Linzen and Baroni, 2021; Peng et al. 2023). These models have been developed by AI specialists and computer programmers using computer languages like Python and C++. They can learn any linguistic phenomenon, be it phonological, morphological, syntactic and/or semantic, provided that they are trained on sufficient and efficient data (Gulordava et al. 2018; Linzen and Baroni 2021; McShane and Nirenburg 2021). They have generative nature/working mechanism like, say, $C_{HL}$ (McShane and Nirenburg 2021). AI DLLMs can learn, process, and compute any linguistic phenomenon in any language. However, this learnability is conditioned by their being exposed to sufficient and efficient linguistic input required for learning to take place, much the same way a child/adult brain needs sufficient and efficient linguistic input to acquire any natural language spoken around him/her (see e.g. Chomsky 1965, *et seq*; Cook 1983, White 2003; Shormani 2023).

Introducing NNAs and DLLMs perhaps marks the real revolution in AI sphere. DNNs "are mathematical objects that compute functions from one sequence of real numbers to another sequence", by means of "neurons". Linguistically, DNNs "encode words and sentences as vectors … these vectors, which do not bear a transparent relationship to classic linguistic structures, are then transformed through a series of simple arithmetic operations to produce the network's output" (Linzen and Baroni 2021: 196). DNNs dominate NLP works "deriving semantic representations from word co-occurrence statistics" (Pavlick 2022: 447). There are also other types of deep earning networks, viz., recurrent neural networks (RNNs). These networks constitute a mechanism that encodes word sequences, in a left-to-right fashion, "maintaining a single vector, the so-called hidden state, which represents the first *t* words of the sentence" (Linzen and Baroni 2021: 197).

## 3.4. LLMs' language processing/interpreting

The correlation of linguistics and AI results in initiating an area of study called NLP, whose developments have been continued since its inception in 1966. It is a field of computer science and technology, the main aim of which is to make computers generate, process and interpret human language. The ideas and projects by NLP and AI scientists were first crystalized in question-answering systems, machine translation, and man-machine conversation (Kenny 2022). NLP depends on linguistic corpora, which are Monolingual, Bilingual, or Multilingual. It has been involved in developing algorithms and models that enable computers to generate human language in such a way that could be understood by humans. It comprises several applications such as machine translation (MT), thus translating texts and documents from one language into another. Another field is Speech Recognition in which a spoken text/language is converted into a written text/language. It includes several applications like sound-text function in WhatsApp, in which a user can speak and this app converts his/her speech into a written text. NLP is a means of analyzing sentiments. In sentiment analysis, computers can process large language data and interpret emotions, feelings, and opinions. Recently, NLP has been used to develop neural networks and algorithms utilized in agents engineering, launching several conversational agents. These agents are employed in chatbots and virtual assistants that can perform conversations with users in a man-man manner. For example, ChatGPT and BERT can engage with humans in conversations of various topics, generating human-like language

NLP includes also parsing, i.e. processing natural language data and parsing it. Consider (25).

(25) a. I think that you like syntax.

b. [TP [Spec [I ]] [T$^I$ [T [think ]][$_v$P [Spec [I~~]][$v$I [$v^I$ [? ~~think~~ ]][VP [Spec [ ]][V$^I$ [V ~~think~~ ]][CP that you like syntax ]]]]]] c.
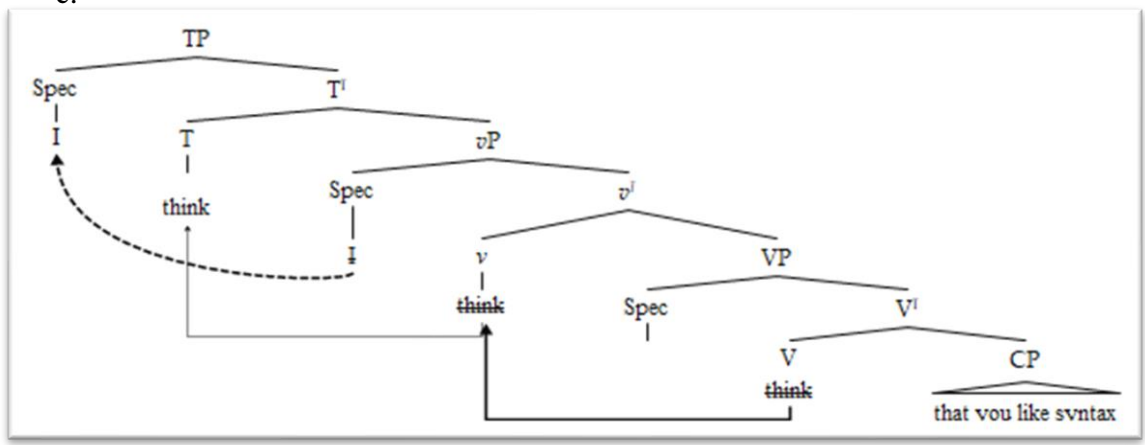

In (25b), the parsing is in the form of bracketing method. In (25c), however, it is in the form of phrase marker (Tree Diagram).

In what follows, we will just focus on the syntactic and semantic phenomena used by AI learning models. As for syntax, DNNs were trained on several and various syntactic phenomena including, for instance, English filler-gap dependencies. In this experiment, DNNs were trained to identify the next n-gram, without any specific supervision on this construction (Gulordava et al. 2018). In a filler–gap dependency, a wh-licensor setting up a prediction for a gap-one of the noun phrases in the embedded clause must be omitted (Gulordava et al. 2018: 200):

(26) a. I know that you insulted your aunt yesterday.   (no wh-licensor, no gap)
     b. *I know who you insulted your aunt yesterday. (wh-licensor, no gap)

In examples (26), (26a) is syntactically well-formed while (26b) is not, and its ungrammaticality lies in the fact that using *who* entails the omission of the NP *your aunt*.

Long-distance agreement is another syntactic phenomenon in which a constituent α agrees with a constituent β, where α and β are far from each other as in (27b):

(27) a. The vibrant linguist is Noam Chomsky.

   b. The vibrant linguist *who wrote several syntax books* is/*are Noam Chomsky.

In (27a), the NP *the vibrant linguist* agrees with the verb *is*, and they are adjacent (not far from each other). However, though they are not adjacent in (27b), they agree also in all φ-features (person, number and gender). The verb *is* agrees with the subject *the vibrant linguist* though there are five words, or otherwise the embedded clause, *who wrote several syntax books*, is between both constituents. Here the words *who wrote several syntax books* between the head of NP *the vibrant linguist*, which is *linguist*, and the verb *is*, are called attractors because they intervene between the subject *linguists* and the verb *is*.

In fact, LDA has received much research in syntactic inquiry cross-linguistically (see e.g. Polinsky and Potsdam 2001; Chomsky 2001, 2005, 2008, Ackema et al. 2006; Koeneman and Zeijlstra 2014; Rouveret 2008; Shormani 2017, 2024). It has also received much interest in AI, specifically deep learning language models (Linzen et al. 2016; Gulordava et al. 2018; Linzen and Baroni 2021; Thrush et al. 2020). In these studies, deep learning models such as DNNs, RNNs were trained on data involving LDA, and the performance of these deep learning models was considerably high, scoring high levels of accuracy, and sometimes even surpasses humans (Gulordava et al. 2018; Kung et al. 2023). In Gulordava et al.'s (2018) experiment, for example, the accuracy rate was 82%. However, the accuracy rate of DNNs changes the more attractors we introduce. Put differently, DNNs were unable to predict LDA beyond 5-grams.

As for semantics, several studies have tackled the semantic-AI interaction, accelerating the semantic bases in AI and how linguistics, in general, contributes to the advancement of AI. The semantic bases in AI have been tackled in relation to several semantic phenomena. For example, Ettinger (2020) studies how AI language models can be trained on argument structure, a semantic structure involving thematic roles such as agent and patient. In particular, Ettinger tested whether BERT (=Bidirectional Encoder Representations from Transformers) can identify argument structure and the semantic role an NP can carry, differentiating between, for instance, agent and patient. Table 1 presents examples of the argument structure data used for training:

**Table 1: Examples of argument structure (from Ettinger 2020: 38)**

| Context 1 | Compl | Context 2 | Match | Mismatch |
|---|---|---|---|---|
| *The restaurant owner forgot which customer the waitress had ____* | *served* | *A robin is a ____* | *bird* | *tree* |
| *The restaurant owner forgot which waitress the customer had ____* | *served* | *A robin is not a ____* | *bird* | *tree* |

The author utilized psycholinguistic stimuli to enhance the training process, and the performance of BERT in this experiment was good enough, i.e. 86% accuracy. Moreover, commonsense knowledge, as a semantic phenomenon, utilized in the training of neural models, was examined by Ettinger (2020). Ettinger trained BERT to recognize hyponym–hypernym relations. The prompts used include *A robin is a [MASK]*, and BERT's performance was considerably high. For example, deciding whether *bird* or *tree* (Table 1), BERT performance was 100%. Additionally, Li et al. (2021) conducted a study, uncovering the implicit representations of meaning in neural language models. They found that dynamic representations of meaning and implicit simulation support prediction in pre-trained neural language models. The ability of BERT to identify novel verb was examined by Thrush et al. (2020). They selected a subclass of verbs based on their selectional restrictions and subcategorization restrictions and trained BERT to do certain tasks.

Another semantic phenomenon on which LLMs were trained is compositionality. The term *compositionality* simply means the meaning of a phrase/sentence is composed of the meaning of words involved and the syntactic/discourse context in which this phrase/sentence is used. It is a property of human language, constraining "the relation between form and meaning" (Everaert et al. 2015: 731). A further semantic phenomenon on which ANNs were trained is systematicity. Systematicity could be defined as "the ability to produce/understand some utterances is intrinsically connected to the ability to produce/understand certain others" (Fodor and Pylyshyn 1988: 37). Thus, if an ANN model can understand the sentence: *Ali loves Alia*, it is expected that it understands the sentence *Alia loves Ali*. Other semantic phenomena that ANNs have been trained on include phrase representations (Shwartz and Dagan 2019), compositionality of negation (Ettinger et al. 2018), polysemy and composition (Mandera et al. 2017), among many other semantic phenomena and scholars. There are studies that tackle linguistics modules interfaces including syntax-semantics interface (Baroni and Lenci 2010), morphology-semantics interface (Marelli and Baroni 2015).

Thus, the syntactic or semantic competence learned by any language model, involving DNNs or RNNs, is a result of training it on massive amounts of data, having the ability to learn in the same way humans acquire language provided that they are exposed to sufficient and efficient linguistic input necessary for language acquisition to take place (Cook 1983; Chomsky 1957, *et seq*; White 2003; Shormani 2023). There is also much involvement of generic architectural properties and features in the same way hierarchical structures including Tree markers or Phrase markers represent how a piece of language is derived and processed, and the mental properties and capacities involved in processing it (Chomsky 1957, 1965, 2013).

## 4. Conclusion

In this article, we provided an account of how generative linguistics contributes to AI, and how GL ideas, conceptions, notions are leveraged in AI since Chomsky's (1956, 1958) and Newell and Simon's (1956) research. Chomsky started working on these things when he was funded by US army and private institutions to develop computer programs that can do human-like tasks. LLMs' working mechanisms to a great extent mirror Chomsky's generative approach which is built on rule and pattern phenomena. The way LLMs learn linguistic phenomena such as LDA, negation, polysemy, compositionality, and compositionality of negation, mirrors the way human infants acquire their L1, and adults as in the case of L2 acquisition (see e.g. O'Grady 2005; O'Grady and Lee 2023). Additionally, the way these things process, compute and interpret these linguistic phenomena also mirrors how $C_{HL}$ processes, computes and interprets language produced by, or spoken to us. These points mark where GL and AI converge.

However, there are some pints where GL and AI diverge. We will address only two: the first has to do with the type of language input. While human brain deals with and handles verbal language, LLMs deal with and handle written language. The first language infants acquire is spoken, they are exposed to spoken input from the people around in an interactive communicative way (see e.g. Beuls and Eecke 2024). However, LLMs learn language from massive and systematized language input. After being trained on huge amounts of written language, LLMs can generate human-like language. The second point of divergence between AI and GL concerns the nature of language input. A human brain gets the linguistic input from "much noise", what is referred to by Chomsky as "just a mess", but an LLM gets its input from statistically systematic data, usually in the form of corpora like BooksCorpus. The former is referred to as the "poverty of stimulus", a human brain comes to life as a "black box", or "tabula rasa", which is not the case with LLMs. Human child acquires his/her language unconsciously and automatically when his/her brain is not mature enough, constructing a stupendously sophisticated grammar of the language spoken around, a grammar consisting of biologically endowed principles and parameters. The grammar constructed is kind of innately, genetically predisposed installation as "anexpression", an "operating system" in our brains endowing us "with the capacity to generate complex sentences and long trains of thought" (Chomsky et al. 2023), which is not the case with LLMs. However, this view might change if one considers symbolic or neuro-symbolic AI, which is a branch of AI employing Chomsky's ideas (see e.g. Maruyama 2021). Maruyama proposes an integrating approach to AI of both Symbolic and statistical theories of AI to overcome the statistical problems of LLMs including explainability and ethical problems.

However, we cannot expect "statistical" behaviorism-based models with behaviorist simple ideas, thinking, reasoning, rationales, etc., coming out of "animal world" to be the basis of such huge and promising progress taking place in AI technology. This said, we cannot even compare Chomsky's revolutionized and sophisticated ideas, thoughts and theorems to that of behaviorism. This implies that the only source of such huge progress in AI is attributed to GL, and that GL has contributed much to AI. Today, Chomsky's biological and cognitive approach marks the conclusive and outstanding triumph over Skinner's behaviorism. The former affects several other scientific disciplines, in what is known as the "cognitive revolution" (Katz 2012). It has been stated long back that it is not only AI that has been influenced by Chomsky's ideas, thoughts, reasoning, and outstanding research, but some other fields of study have been influenced including "sociolinguistics, ethnomethodology, anthropology, psychoanalysis, the psychology of language, pragmatics, and philosophy of action" (Goodwin and Hein 1982: 249). What Chomsky and his school aim at is seize language acquisition as the central problem to linguistic theory, characterize the uniqueness of human brain, or what makes it so special, be it in UG, FL, or so. This actually unveils an Ayat in the Holy Qur'an: "And He (God) teaches Adam all the names"[2], where "names" sometimes refer to "languages". God "installs" this "language" in our brain, not only to communicate, but more importantly, to think. Chomsky by and large aims to provide evidence "that language is primarily an instrument of thought, not of communication" (Chomsky 2016: 13). Thus, whatever success and expectations AI reaches, it will never reach the human brain State or working mechanism; it is a species-specific property, which makes it so special. It is not Isaac Asimov's 'robot', Joseph Weizenbaum's 'ELIZA', nor is it OpenAI's 'ChatGPT', 'BERT', or whatever. In critiquing Chomsky, Hinton seems not to understand the powers of UG, TGG, $C_{HL}$, and the pivotal aspects of the "Chomsky School".

A final remark, isn't this all employed in AI? Doesn't this all belong to the "Chomsky School". Aren't Chomsky's ideas, conceptions of human brain, UG, FL, $C_{HL}$ working mechanism, language acquisition models, processing/computational models, and so on utilized in AI? And isn't this a truism? The answer to all these questions is, indeed, *YES*.

---

[2] Surat AlBaqara, Ayat (31).